\documentclass{article}

\usepackage{arxiv}

\usepackage[utf8]{inputenc} 
\usepackage[T1]{fontenc}    
\usepackage{hyperref}       
\usepackage{url}            
\usepackage{booktabs}       
\usepackage{amsfonts}       
\usepackage{nicefrac}       
\usepackage{microtype}      
\usepackage{lipsum}		
\usepackage{graphicx}
\usepackage{natbib}
\usepackage{doi}

\title{Evaluating a Synthetic Image Dataset Generated with Stable Diffusion}

\date{October 28, 2022}	

\author{ \href{https://orcid.org/0000-0003-1646-0514}{\includegraphics[scale=0.06]{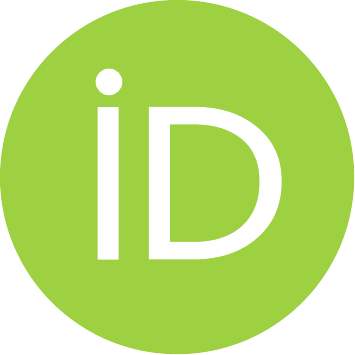}\hspace{1mm}Andreas Stöckl} \\
	Department Digital Media, \\
	University of Applied Sciences Upper Austria, \\
	Hagenberg, Austria \\
	\texttt{andreas.stoeckl@fh-hagenberg.at} \\
}

\renewcommand{\shorttitle}{Evaluating a Syntectic Image Dataset}

\begin{document}
\maketitle

\begin{abstract}
	We generate synthetic images with the "Stable Diffusion" image generation model using the Wordnet taxonomy and the definitions of concepts it contains. This synthetic image database can be used as training data for data augmentation in machine learning applications, and it is used to investigate the capabilities of the Stable Diffusion model.
	
	Analyses show that Stable Diffusion can produce correct images for a large number of concepts, but also a large variety of different representations. The results show differences depending on the test concepts considered and problems with very specific concepts. These evaluations were performed using a vision transformer model for image classification.
	
\end{abstract}


\section{Introduction}
Current models for synthetic image generation can not only produce very realistic-looking images, but also deal with a large number of different objects. In this paper we use the example of the model "Stable Diffusion" to investigate which objects and types are represented so realistically that a subsequent image classification assigns them correctly. This will give us an estimate of the models potential with regard to realistic representation.

Pre-trained models, such as the one we present, also form the basis for further finetuning to specific objects, as described in \cite{ruiz2022dreambooth}, and only need a few images of the object. The prerequisite, however, is that the basic model can cope with the desired objects and object classes.

With "Stable Diffusion" we use a current model for image generation to create an artificially generated data set for training image processing systems. We then evaluate the model using image classification. For the categorisation of classes, we use the same approach as ImageNet \cite{imagenet_cvpr09}, which uses nouns from Wordnet \cite{miller1995wordnet}.

For the subset of our dataset that corresponds to the classes in the ImageNet Large Scale Visual Recognition Challenge (ILSVRC) \cite{russakovsky2015imagenet}, we test with an actual image classification method to see how well our synthetic images can be classified. For this, we use the Pytorch implementation of the vision transformer vit\_h\_14 model from cite{dosovitskiy2020image}. Which has a top 1 accuracy of 88.55\% and a top 5 accuracy of 98.69\% on the real Imagenet data.

This synthetic data is also a good way to improve the diversity of data in a supervised learning setting. They help to get more data without the time-consuming labelling process. Synthetic data can also be seen as the logical extension of data augmentation (e.g. \cite{shorten2019survey}, \cite{perez2017effectiveness}, \cite{inoue2018data}), which is standard in image processing. \cite{seib2020mixing} gives an overview of different approaches to enriching real data with synthetic data.

\section{Related Work}

The "Stable Diffusion" model \cite{Rombach_2022_CVPR} we use is the latest representative of diffusion models for image generation. The basis of these models was the work of \cite{sohl2015deep} which was improved in \cite{song2019generative} and \cite{ho2020denoising}. Important and well-known other implementations are "Google Imagen" \cite{saharia2022photorealistic}, "GLIDE" \cite{nichol2021glide}, "ERNIE-ViLG" \cite{zhang2021ernie}, "DALLE" \cite{ramesh2021zero} "Dalle-Mini", \cite{daymadallemini2021} and "DALLE 2" \cite{ramesh2022hierarchical}. Examples of other image generators are "Midjourney" (\url{https://www.midjourney.com/}) and "Google Parti" \cite{yu2022scaling}.

\cite{borji2022good} investigates how well images generated by DALL-E 2 and Midjourney perform in object recognition and visual question answering (VQA) tasks and compares the results with those on real Imagenet images. The results for the synthetic images are significantly worse and the authors conclude that "deep models struggle to understand the generated content". 

In \cite{borji2022generated}, Stable Diffusion, Midjourney, and DALL-E 2 are examined to see how well they perform in generating faces. They find that Stable Diffusion generates better faces than the other systems. Furthermore, a dataset containing images of faces is provided.

For the training of object recognition methods \cite{hinterstoisser2018pre}, \cite{rozantsev2015rendering}, \cite{rajpura2017object} and segmentation \cite{ros2016synthia}, \cite{McCormac_2017_ICCV}, the use of different synthetic image data has been common for some time. Here, the use of synthetic image generators, as described above, offers a variety of further possibilities.

\begin{figure}
	\centering
	\includegraphics[width=\columnwidth]{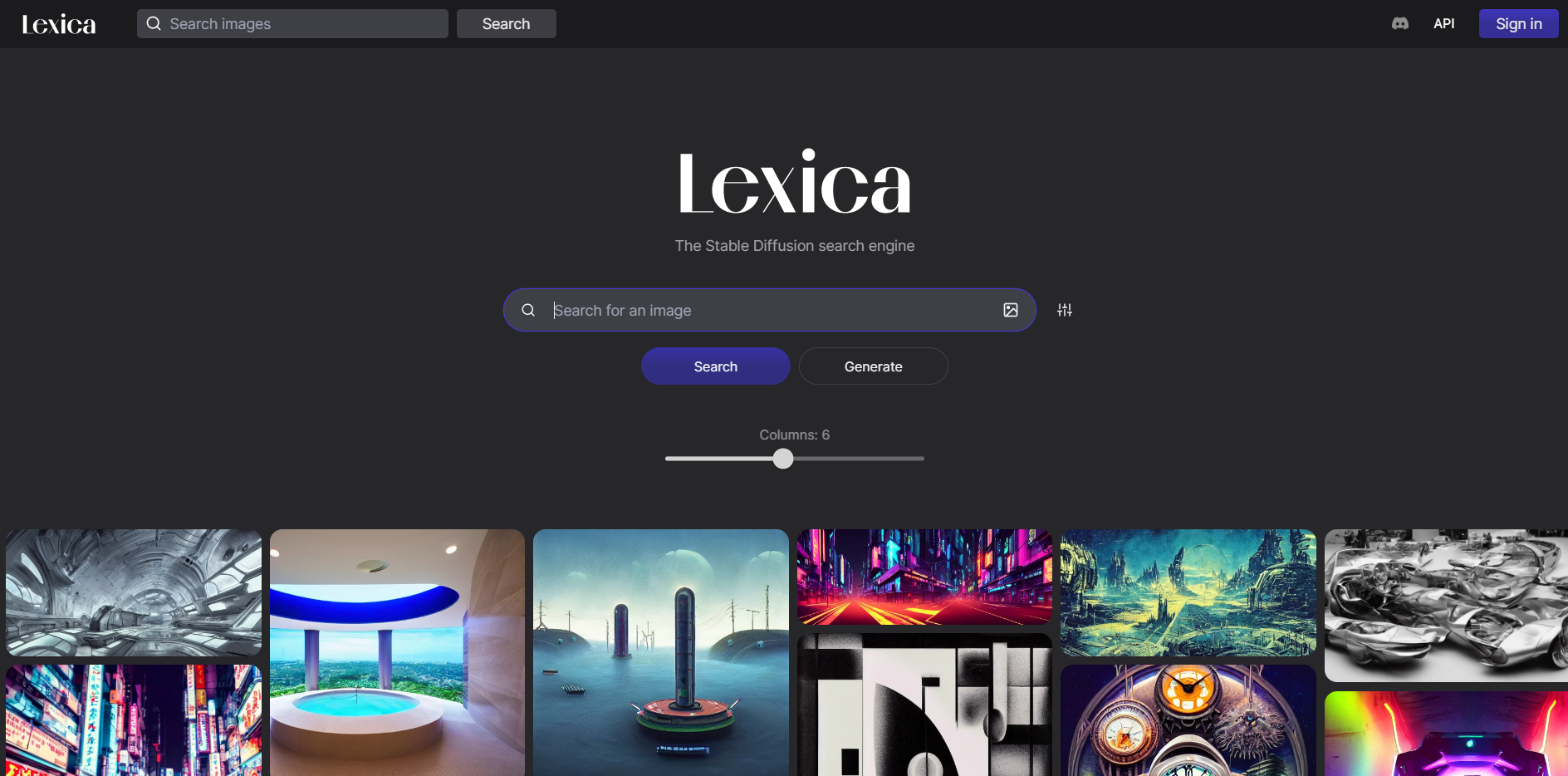}
	\caption{Lexica.art a Synthetic Image Search}
	\label{fig:lexica}
\end{figure}

A project that provides access to synthetic image data generated with Stable Diffusion is "Lexica" (Fig. \ref{fig:lexica} - \url{https://lexica.art/}). It is a search engine that returns results for a term from over 10 million images. However, the entire database cannot be downloaded here, and there is no categorisation.

A large database of 2 million images, which can also be downloaded and used as open source, is offered and described in \cite{wang2022diffusiondb}. Besides the images the dataset 'DiffusionDB' also contains the text prompt used to generate each image, as is the case in our collection. Since this database consists of images that were created by many different users and settings in Stable Diffusion, in contrast to ours, these settings are also stored for each image. 

The data collection was created by the authors crawling the Discord server of Stable Diffusion and extracting the images including the prompt. Unlike our collection, this does not result in systematic coverage of the wide range of possible concepts, but rather a bias towards the applications that were of interest to the testing community. It also lacks the hierarchy that we have available through the use of Wordnet, and use for analysis. The potential applications of "DiffusionDB" that are discussed focus on prompt engineering and explanations, and studies of deepfakes.

\section{Generation of the Data}

As a basis for image generation, we use the "Stable Diffusion" 1.4 model with their Huggingface Diffusers library \cite{von-platen-etal-2022-diffusers} implementation. This is a model that allows images to be created and modified based on text prompts. It is a latent diffusion model \cite{Rombach_2022_CVPR} trained on a subset (LAION-Aesthetics) of the LION5B text to image dataset, and uses the pretained text encoder CLIP ViT-L/14 \cite{radford2021learning} to encode the text prompts, as proposed by Imagen \cite{saharia2022photorealistic}.

\begin{figure}
	\centering
	\includegraphics[width=300pt]{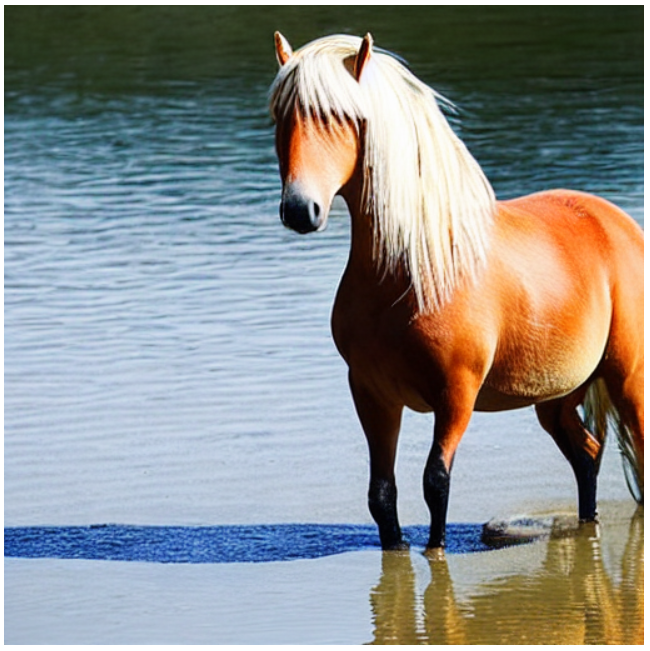}
	\caption{Image for the text "haflinger horse with short legs standing in water"}
	\label{fig:horse}
\end{figure}

Figure \ref{fig:horse} shows an example of an image generated from the text prompt "haflinger horse with short legs standing in water". The example shows that the generator model can represent different concepts with varying attributes and can also combine them in a setting. We now want to create a dataset that contains images of a variety of different concepts in order to evaluate the results.

For text input, we use the information contained in "Wordnet" \cite{miller1995wordnet}. Wordnet organises concepts into so-called "synsets", which correspond to a meaning of one or more words with the same meaning. A word with different meanings can thus belong to several synsets. For example, the word "apple" has the meanings of a fruit and a computer brand, each with a synset for these concepts.

In addition to the name, further information is contained for each synset, such as a unique wordnet number and a definition. A directed graph spans between the synsets, which represents the relationships "hypernyms" (a word with a broad meaning constituting a category into which words with more specific meanings fall) and "hyponyms" (a word of more specific meaning than a general or superordinate term applicable to it), and thus makes the hierarchical relationships derivable by being able to output superordinate terms and subordinate terms of a concept.

Starting from the Wordnet synset 'object.n.01', we build a list of 26,204 synsets of nouns by recursively calling the "hyponyms". For each of these nouns, we use the description of the synsets in Wordnet for the text prompts of the image generator. For each synset, 10 images are generated and stored under the name of the synset with the number appended. This results in a total of 262,040 images for our dataset. The default settings for Stable Diffusion are 512x512 pixels, 50 inference steps, Guidance Scale 7.5 and PLMS sampling \cite{liu2022pseudo}. On an RTX3090 GPU, it took about 6 seconds to create an image. This resulted in a total time of more than 18 days for the creation of the whole dataset.

An example of such a prompt is: (synset for dogs)

“a member of the genus Canis (probably descended from the common wolf) that has been domesticated by man since prehistoric times; occurs in many breeds”

Together with the 10 images per synset, a text file is saved that contains the name of the synset (e.g. "dog.n.01") and the wordnet number (e.g. "n12345678") in addition to the prompt used. The dataset can be downloaded from Kaggle \url{https://www.kaggle.com/datasets/astoeckl/stable-diffusion-wordnet-dataset}.

\section{Results}
\label{sec:results}

\begin{figure}
	\centering
	\includegraphics[width=\columnwidth]{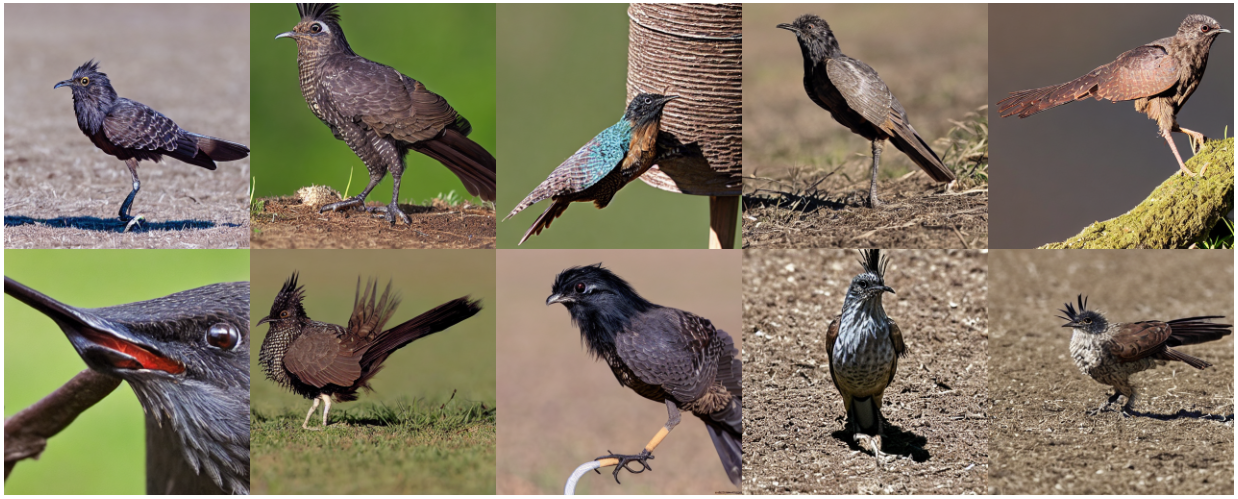}
	\caption{Generated Images for "Coucal"}
	\label{fig:coucal}
\end{figure}

\begin{figure}
	\centering
	\includegraphics[width=\columnwidth]{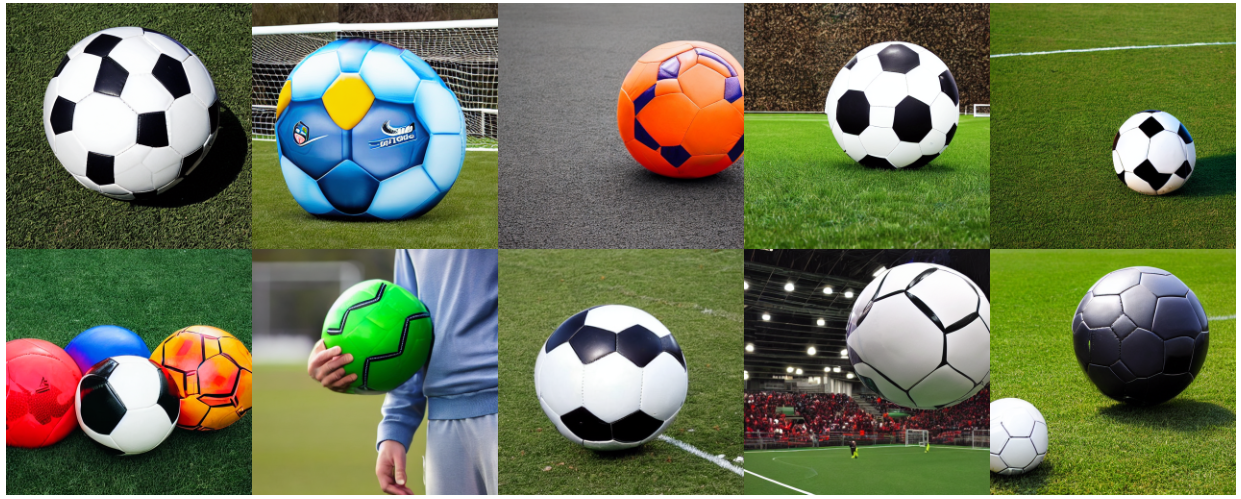}
	\caption{Generated Images for "Soccer Ball"}
	\label{fig:soccer}
\end{figure}

First, let's look at some examples of generated images. Figure \ref{fig:coucal} shows the images generated for the term "Coucal", Figure \ref{fig:soccer} for the term "Soccer Ball". It shows on the one hand that very realistic photos were created, and on the other hand a large variety in the representation.

\begin{figure}
	\centering
	\includegraphics[width=\columnwidth]{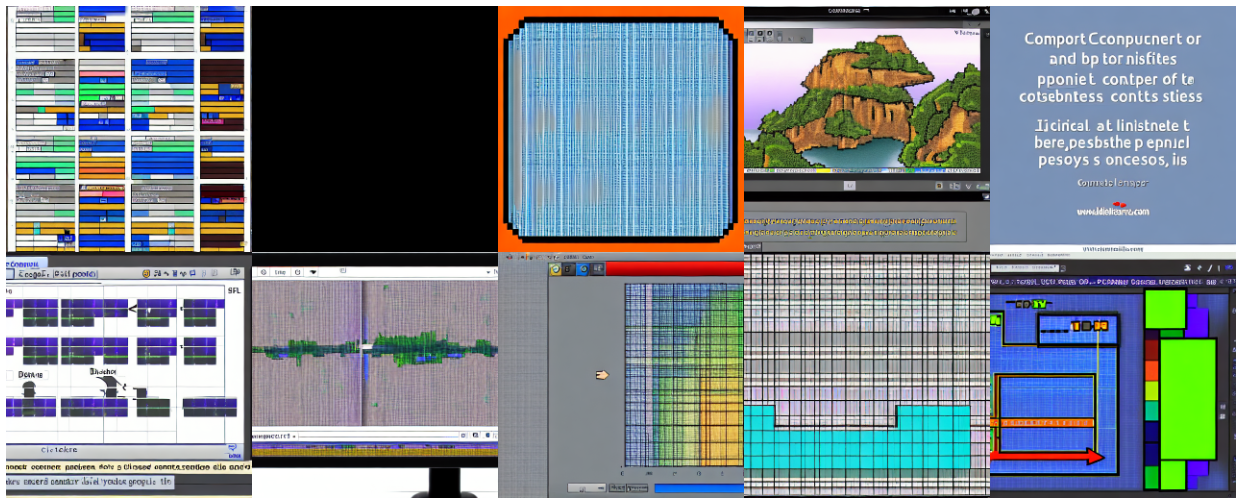}
	\caption{Generated Images for "Frame Buffer"}
	\label{fig:frame_buffer}
\end{figure}

Figure \ref{fig:frame_buffer} shows the attempt with the abstract term "Frame Buffer". Here, the model naturally finds it difficult to generate suitable images.

\subsection{NSFW Filter}

Stable Diffusion has a safety filter that is supposed to prevent the generation of explicit images. Unfortunately, the functionality of the filter is obfuscated and poorly documented. In \cite{rando2022red}, it was found that while it aims to prevent sexual content, it ignores violence, gory scenes, and other similarly disturbing content.

\begin{figure}
	\centering
	\includegraphics[width=\columnwidth]{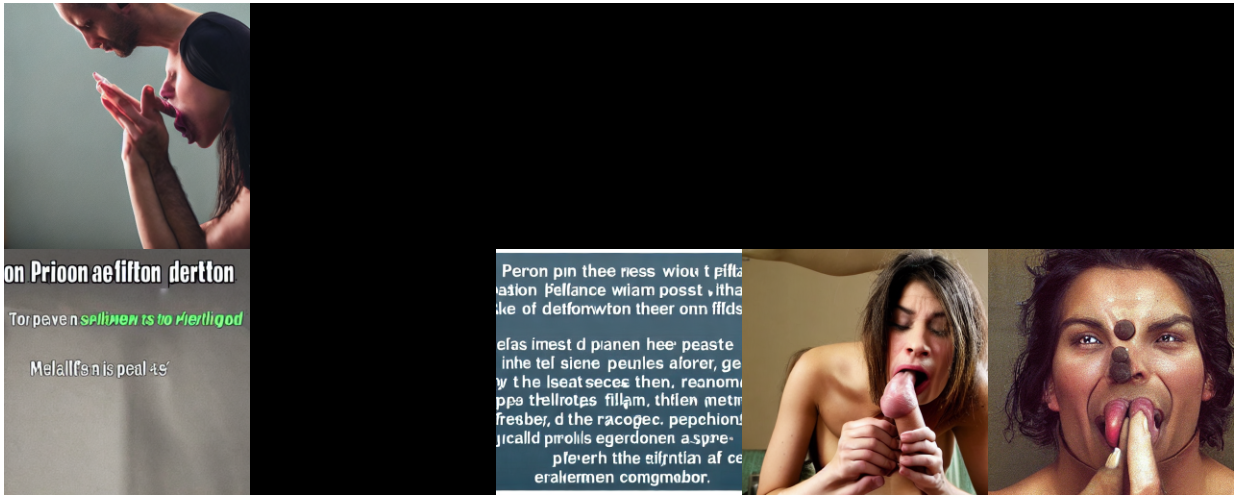}
	\caption{Generated Images for "Cocksucker"}
	\label{fig:sucker}
\end{figure}

Our tests with sexual content have shown that the filter does not work reliably here either (see Figure \ref{fig:sucker} for "Cocksucker"). A black image indicates that the filter has suppressed the output.

An example of filters that have triggered incorrectly is shown in Figure [5] for the term "System Clock".

\begin{figure}
	\centering
	\includegraphics[width=\columnwidth]{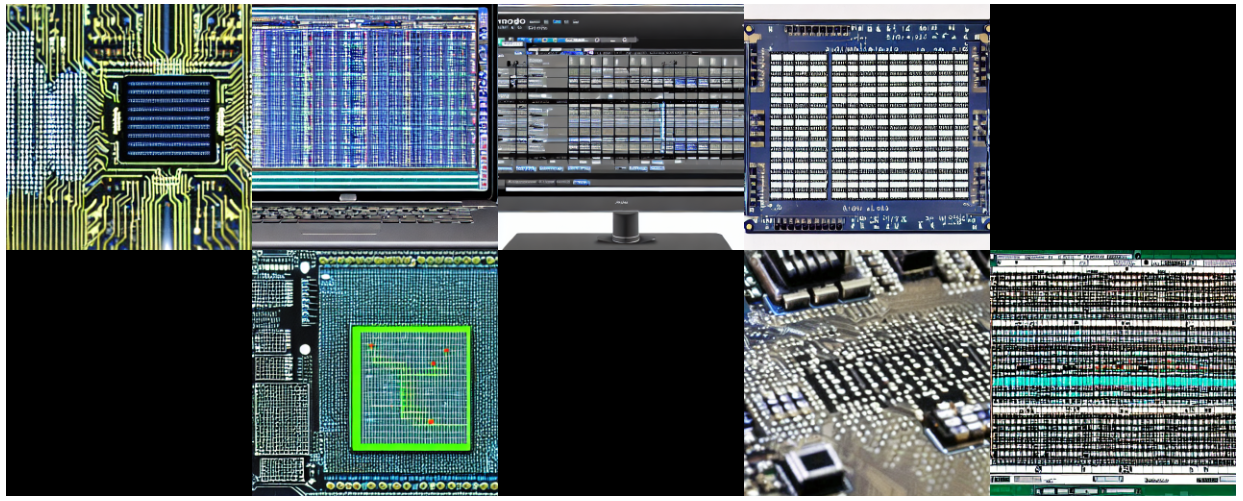}
	\caption{Generated Images for "System Clock"}
	\label{fig:system_clock}
\end{figure}

We examine which classes trigger the filter and therefore generate black images. In total, 4620 black images were generated. This is a percentage of 1.76\% over all images.

\subsection{Classification with Vision Transformer}

We not only to look at and evaluate individual images, but to perform systematic evaluations for a subset of our dataset that is included in the ImageNet Large Scale Visual Recognition Challenge (ILSVRC) \cite{russakovsky2015imagenet}. That is 861 classes. We use the Pytorch implementation of the vision transformer vit\_h\_14 model from \cite{dosovitskiy2020image}, which has a top 1 accuracy of 88.55\% and a top 5 accuracy of 98.69\% on the Imagenet data, to verify that the generated images can be correctly classified.

\begin{figure}
	\centering
	\includegraphics[width=\columnwidth]{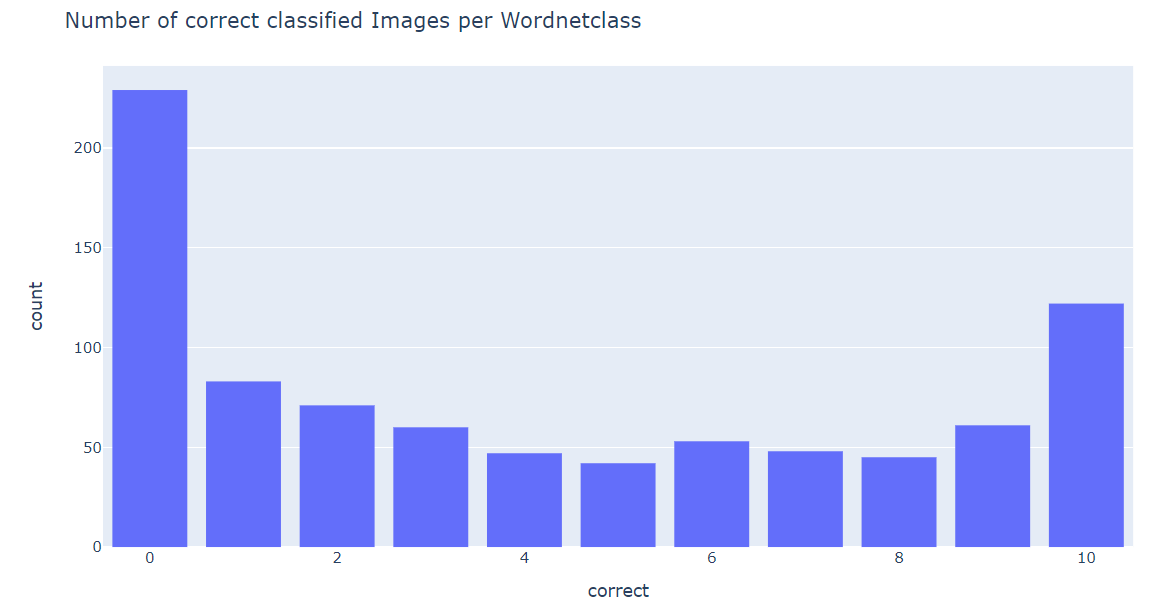}
	\caption{Number of correct classified images per class}
	\label{fig:histogram}
\end{figure}

A review of all 8610 images from the considered subset yields a average correct classification of 4.16 images per class (Maximum 10) with a average standard deviation of 3.74, across all classes. The histogram in Figure \ref{fig:histogram} shows the large spread in the number of correct classifications. The black images generated by the NSFW filter are part of the statistics.

It can be seen that although at least one correctly recognized image was generated for a large part of the classes (73\%), only for 14\% of the classes all 10 images were recognized again. This also reflects the observation made at the beginning of section \ref{sec:results} on the basis of some examples that the generated images of a class differ strongly. This complicates the task for the classification procedure.

In the Wordnet Taxonomy there is a "depth" parameter that specifies how many steps you have to descend from the base class to get to the given class. It is therefore a measure of how specific a class is. We now investigate the dependence of the classification rate on the depth. 

\begin{figure}
	\centering
	\includegraphics[width=\columnwidth]{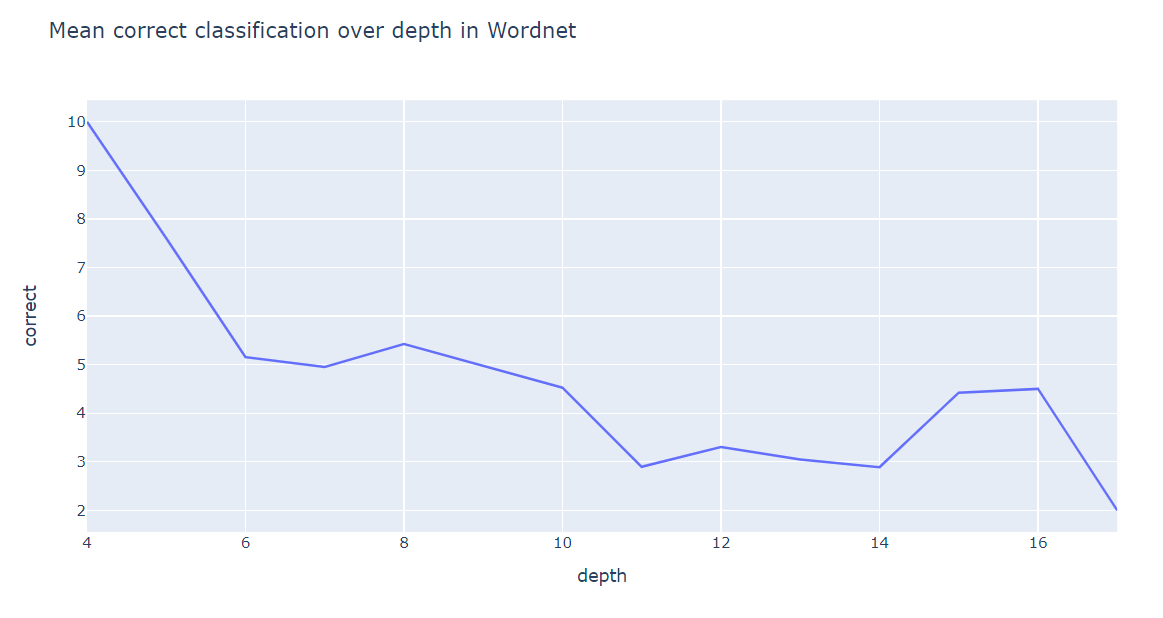}
	\caption{Number of correct classified images over "depth" in wordnet}
	\label{fig:depth2}
\end{figure}

Looking at the mean recognition rate as a function of depth, the picture of Figure \ref{fig:depth2} emerges, indicating a slightly decreasing recognition rate with increasing depth. Generated images of more specific concepts are thus somewhat more difficult to classify correctly.

Let us now consider the recognition rate of some groups of objects. Using the hierarchy of Wordnet, the associated classes were combined for some groups of concepts, and the average recognition rate was determined in each case. Table \ref{tab:tablegroups} shows the results.

\begin{table}
	\caption{Recognition rate of different object classes}
	\centering
	\begin{tabular}{lcll}
		\toprule
		Group & Number of classes & Mean & Std. \\
		\midrule
		Vehicle & 61 & 4.95 & 3.79    \\
	    Animal & 376 & 2.72 & 3.35 \\
        Machine & 14 & 4.14 & 4.26 \\
        Fruit & 8 & 4.5 & 3.55 \\
        Building & 11 & 7.18 & 2.64 \\
        Tool & 13 & 3.85 & 3.46 \\
        \midrule
        Mean & 861 & 4.16 & 3,74 \\
		\bottomrule
	\end{tabular}
	\label{tab:tablegroups}
\end{table}

\begin{figure}
	\centering
	\includegraphics[width=\columnwidth]{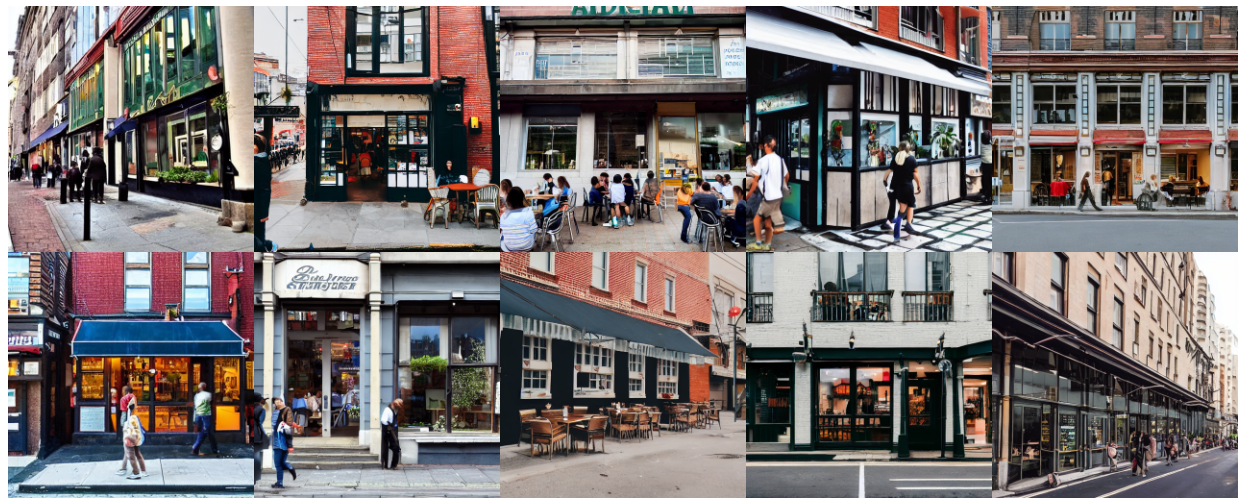}
	\caption{Generated Images for "Restaurant"}
	\label{fig:restaurant}
\end{figure}

\begin{figure}
	\centering
	\includegraphics[width=\columnwidth]{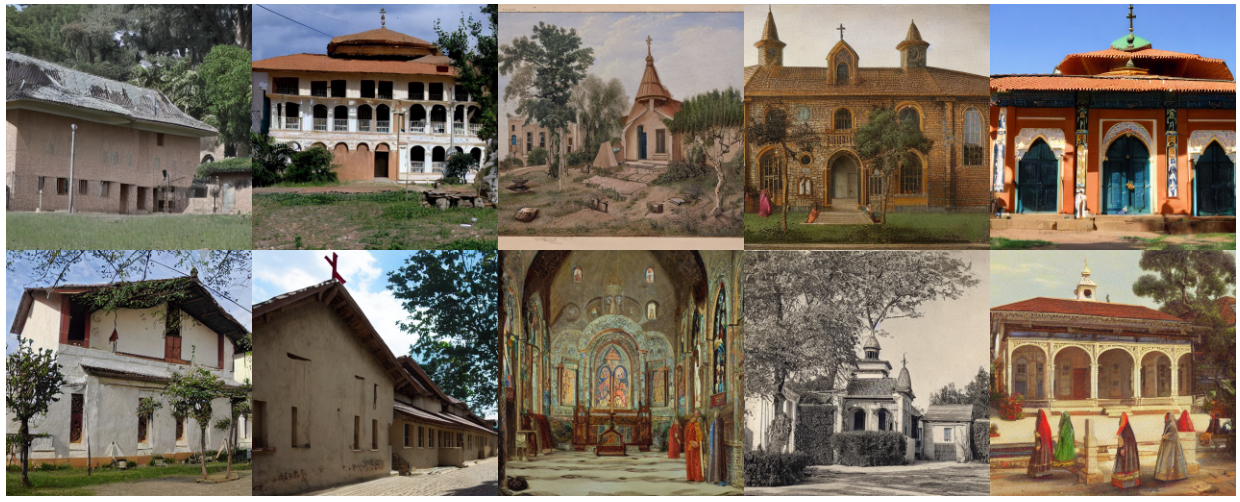}
	\caption{Generated Images for "Monastery"}
	\label{fig:monastary}
\end{figure}

\begin{figure}
	\centering
	\includegraphics[width=\columnwidth]{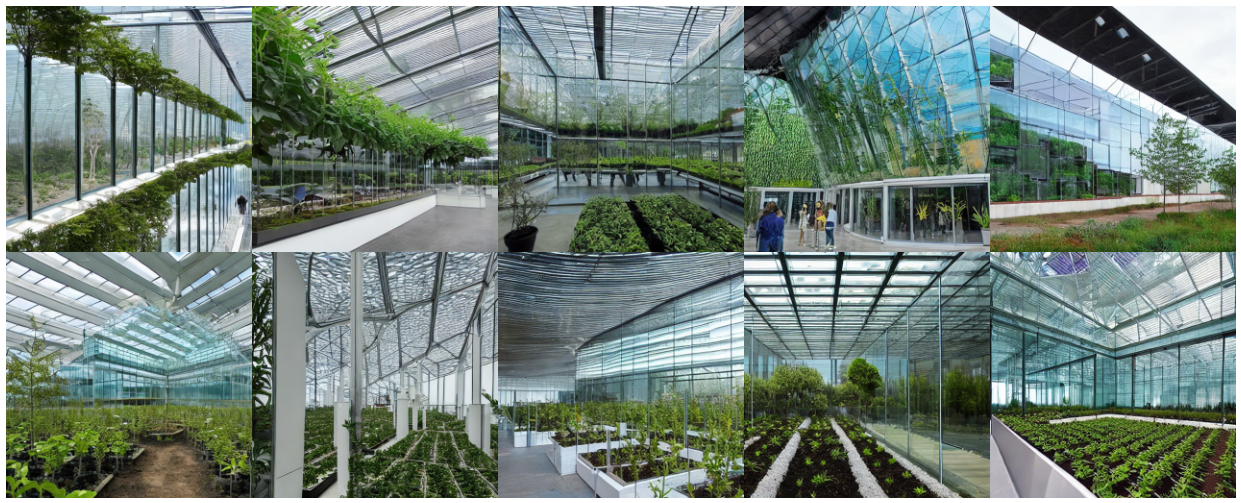}
	\caption{Generated Images for "Greenhouse"}
	\label{fig:greenhouse}
\end{figure}

Remarkable are the good recognition rates for buildings. Fig. \ref{fig:restaurant} and Fig. \ref{fig:monastary} show the images for the term "Restaurant" and "Monastery", where 5 each were correctly classified. Figure \ref{fig:greenhouse} shows the images for "Greenhouse" all 10 of which were correctly recognized.

The "Animal" superclass shows below-average classification rates. If we look a little closer at this group, we see that for 162 animal classes no image was recognized at all. And that the average depth in the Wordnet hierarchy for animals with 12.2 is higher than the overall average of the test classes with 10.4. The test set thus not only contains a particularly large number of classes 376 for animals, but also particularly specific ones. These may have made detection more difficult.

\begin{figure}
	\centering
	\includegraphics[width=\columnwidth]{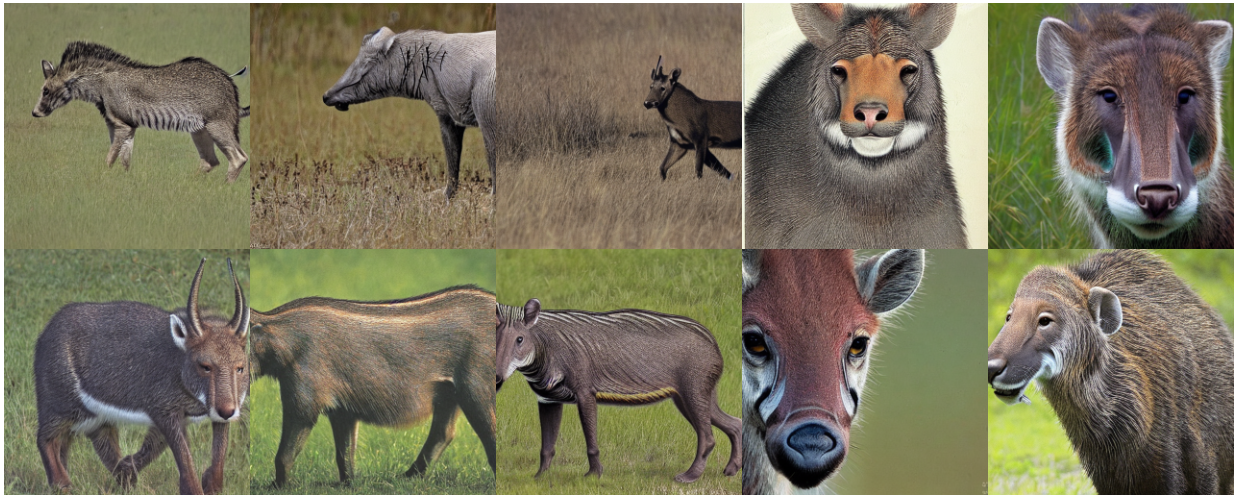}
	\caption{Generated Images for "Black footed ferret"}
	\label{fig:ferret}
\end{figure}

\begin{figure}
	\centering
	\includegraphics[width=\columnwidth]{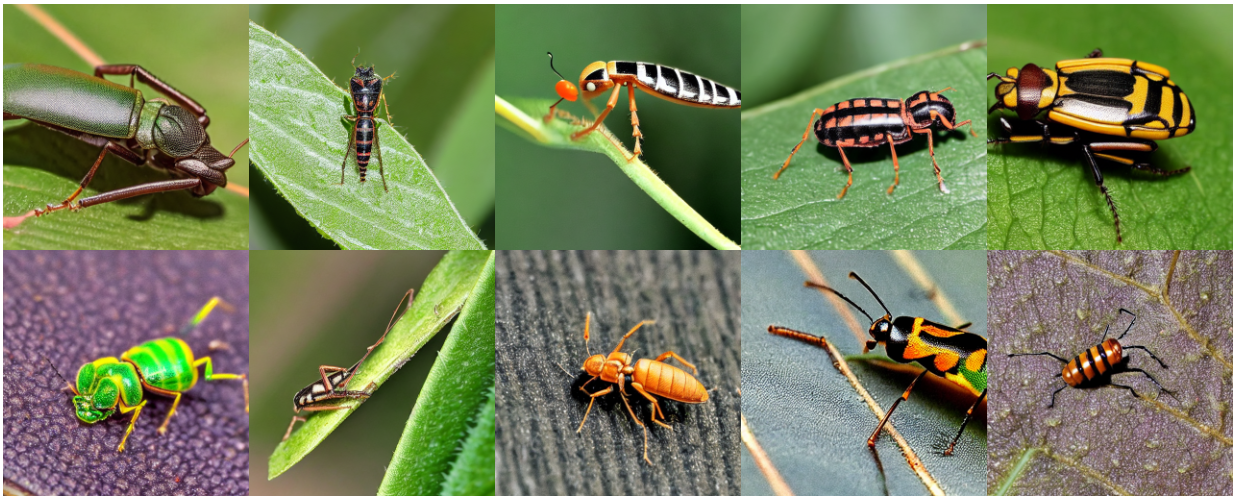}
	\caption{Generated Images for "Leafhopper"}
	\label{fig:leafhopper}
\end{figure}

Looking at individual specific examples, such as Fig. \ref{fig:ferret} (showing an example of the term "black footed ferret") and Fig. \ref{fig:leafhopper} (showing an example of the term "leafhopper"), "Stable Diffusion" obviously reveals significant deficiencies in animal science.

To further overview the results for groups of concepts, we consider some visualizations in the next section.

\subsection{Visualization with Word Embeddings}

To create a "map" of the terms that shows which of the images generated by Stable Diffusion are correctly recognized by the Vision Transformer model, and how good the recognition rate is in each case, we plot the terms by semantic meaning in 2D and color each by subgroup. The size of a circle indicates the number of correctly classified images. To determine the positions on this map, we use word embeddings \cite{mikolov2013efficient} for the names of the order classes. We use the "Fast Text Model" \cite{bojanowski2016enriching} that was pretrained on Google News and Wikipedia data, since it is trained on the subword level, and the embeddings of the terms can be composed from these. This also avoids that our tested terms are not present in the vocabulary. Using "Gensim"  \cite{rehurek2011gensim}, the model was loaded and the 300 dimensional vectors were extracted.

For the two-dimensional representation a dimension reduction is necessary for which we used PCA \cite{jolliffe2016principal} and TSNE \cite{van2008visualizing}. Scikit Learn \cite{pedregosa2011scikit} was used for the computation. For labels consisting of more than one word, the embeddings of the individual words are added to obtain an embedding of the object.

\begin{figure}
	\centering
	\includegraphics[width=\columnwidth]{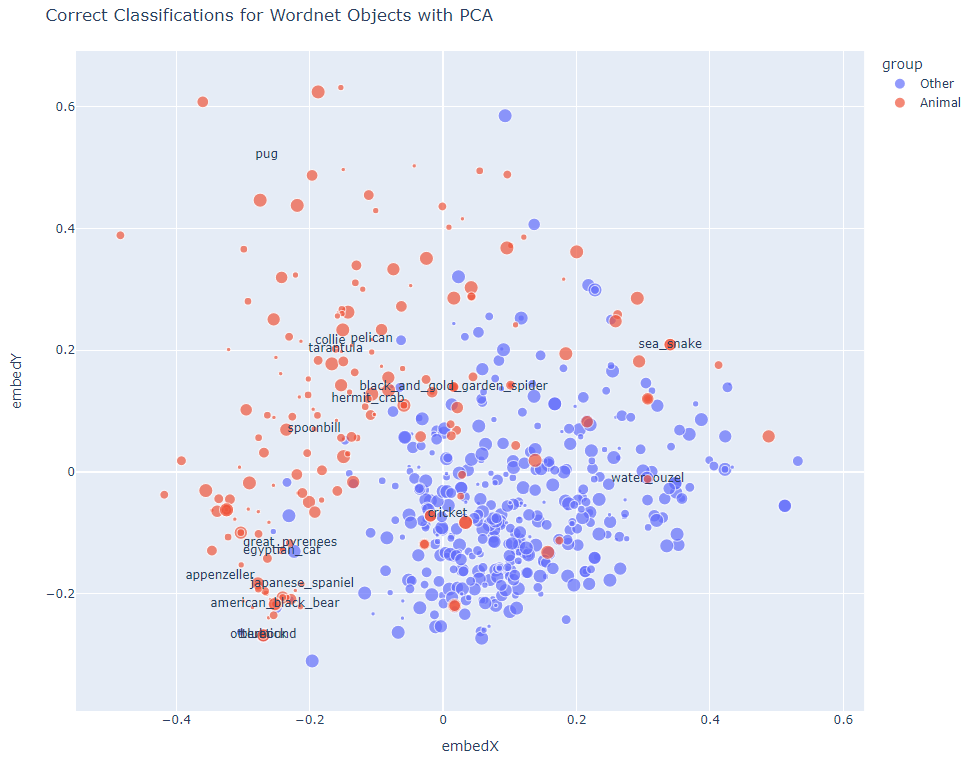}
	\caption{Map of correct classified images for animals with pca}
	\label{fig:animal_pca}
\end{figure}

\begin{figure}
	\centering
	\includegraphics[width=\columnwidth]{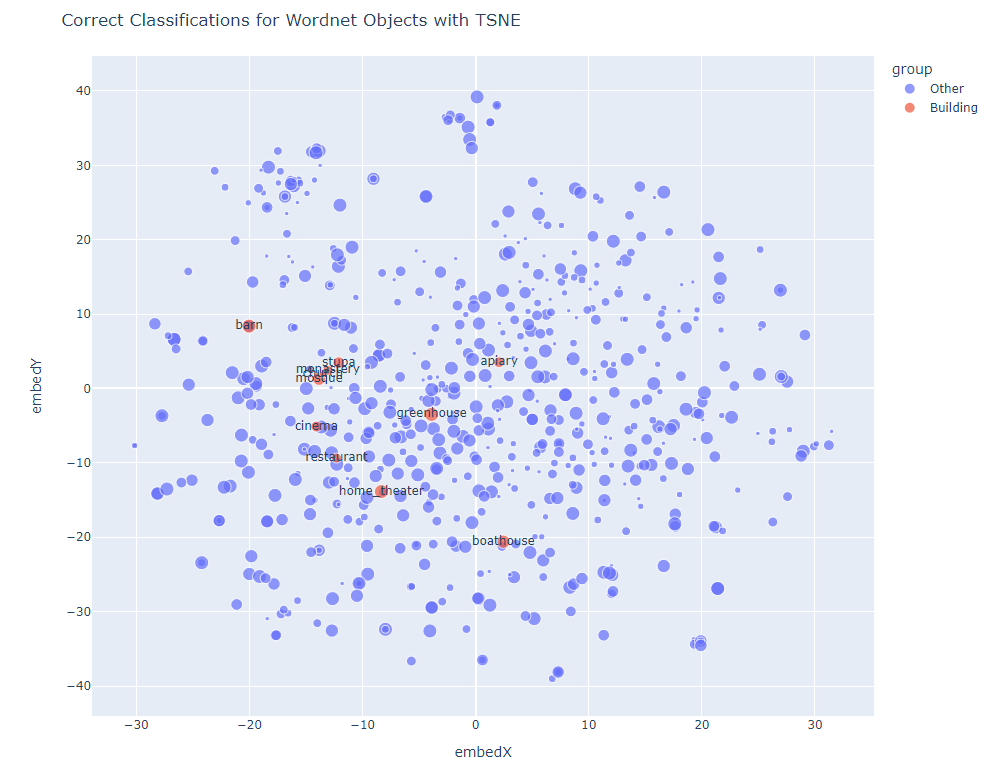}
	\caption{Map of correct classified images for buildings with tsne}
	\label{fig:building_tsne}
\end{figure}

Fig. \ref{fig:animal_pca} for example shows the "map" colored for the superclass "animals" and projected by PCA.  Here, too, the many small red dots corresponding to classes that were not correctly recognized are noticeable, as described in the previous section. Fig. \ref{fig:building_tsne} shows the superclass "Building" projected using TSNE. The different classes are not very well represented here by embedding in a common region. The good classification rate of "Buildings" is shown by the relatively large red circles in the figure.

\section{Summary and Future Work}

Using the Wordnet taxonomy, it was possible to automatically generate synthetic images for a wide spread set of concepts by using the definitions of the concepts as a prompt for Stable Diffusion. 

This data set can now be used as a basis for various image processing applications that use it for data augmentation. It would be interesting to see if image classification or object detection methods can benefit from this data augmentation. It would also be interesting to train an image classification model like Vision Transformer on our synthetic dataset or an even larger one and test it on real data.

A second aspect for which the data set can be useful is to better analyze and understand the generation system "Stable Diffusion". Our first analyses show that Stable Diffusion generated at least 1 correct image for 10 trials for a wide range of concepts (73\%). So for a large part of the world there is information in the system. On the other hand, very different images are generated for one concept, which is useful for a generative system, but this also decreases the recognition rate.

We have also seen that different groups of concepts were "understood" differently, as could be seen for example with very specific animal species. There is plenty of room for further investigation and evaluation here. 

Finally, it has been shown that the system's filter for unwanted content does not work reliably.

\bibliographystyle{unsrtnat}
\bibliography{references}  






\end{document}